\pdfoutput=1

\documentclass{article}

\usepackage{spconf}

\newcommand{\emailUrl}[1]{\href{mailto:#1@apple.com}{#1}}

\title{Server-side Rescoring of Spoken Entity-centric Knowledge Queries\\for Virtual Assistants}

\name{Youyuan Zhang, Sashank Gondala, Thiago Fraga-Silva, Christophe Van Gysel}
\address{Apple \\
         \{\emailUrl{youyuan\_zhang}, \emailUrl{sgondala}, \emailUrl{tfragadasilva}, \emailUrl{cvangysel}\}@apple.com}

\usepackage{amsmath,graphicx}
\usepackage{acronym}

\usepackage[numbers]{natbib}
\usepackage[english]{babel}
\usepackage[inline]{enumitem}
\usepackage{hyperref}
\usepackage{acronym}
\usepackage{amsmath}
\usepackage{booktabs}
\usepackage{siunitx}
\usepackage{xcolor}
\usepackage{makecell}
\usepackage{numprint}
\usepackage{makecell}
\usepackage{subcaption}
\usepackage{graphicx}
\usepackage{soul}

\newacro{AM}{acoustic model}
\newacro{ASR}{Automatic Speech Recognition}
\newacro{BPE}{byte-pair encoding}
\newacro{CNN}{convolutional-based neural network}
\newacro{FOFE}{Fixed-size Ordinally Forgetting Encoding}
\newacro{FST}{Finite State Transducer}
\newacro{KB}{knowledge base}
\newacro{KG}{knowledge graph}
\newacro{LM}{Language Model}
\newacro{LSTM}{Long-Short Term Memory}
\newacro{NER}{Named-Entity Recognition}
\newacro{NNLM}{Neural Network Language Model}
\newacro{NLP}{Natural Language Processing}
\newacro{PPL}{Perplexity}
\newacro{RQ}{Research Question}
\newacro{SP}{SentencePiece}
\newacro{VA}{Virtual Assistant}
\newacro{WER}{Word Error Rate}
\newacro{LLM}{Large Language Model}
\newcommand{\CandidateSetSize}{N}
\newcommand{\NBest}{$\CandidateSetSize{}$-best}
\newcommand{\NBestList}{\NBest{} list}
\newcommand{\NGram}{$N$-gram}

\begin{document}

\maketitle
\begin{abstract}
On-device \acp{VA} powered by \ac{ASR} require effective knowledge integration for the challenging entity-rich query recognition.
In this paper, we conduct an empirical study of modeling strategies for server-side rescoring of spoken information domain queries using various categories of \acp{LM} (\NGram{} word \acp{LM}, sub-word neural LMs). 
We investigate the combination of on-device and server-side signals, and 
demonstrate significant WER improvements of 23\%-35\% on various entity-centric query subpopulations
by integrating various server-side LMs compared to performing \ac{ASR} on-device only.
We also perform a comparison between LMs trained on domain data and a GPT-3 variant offered by OpenAI as a baseline.
Furthermore, we also show that model fusion of multiple server-side LMs trained from scratch most effectively combines complementary strengths of each model and integrates knowledge learned from domain-specific data to a VA \ac{ASR} system.
\end{abstract}

\acresetall

\acresetall

\section{Introduction}
\label{section:introduction}

\newcommand{\LMCategory}[1]{(C{#1})}

\newcommand{\NGramCategory}[1][]{$\mathit{{NGram}}$}
\newcommand{\NNLMCategory}[1][]{$\mathit{{NNLM}}$}
\newcommand{\LLMCategory}[1][]{$\mathit{{LLM}}$}

\Acp{VA} are rapidly gaining popularity \cite{Juniper2019popularity,Statista2022popularity} as they assist users with various tasks \cite{Maarek2019alexa}.
Voice commands issued by \ac{VA} users are recognized using \ac{ASR}, a critical component of any \ac{VA} system. The \ac{VA} \ac{ASR} component takes as input user-spoken audio and generates a ranked list of $\CandidateSetSize{}$ transcription hypotheses, hereafter the \NBestList{}.
A primary challenge of a \ac{VA} \ac{ASR} system is that queries are entity-rich and heavily centered around complex information domains which are usually present in
server-side knowledge bases. 
Here, we refer to any such query that may benefit from a knowledge base as \emph{information domain query}.
For example, consider a query - \emph{``play Red Smoke by The Reytons"} that instructs the VA to play a song. If the corresponding song (i.e., in the previous example, \emph{Red Smoke}) is not available within the user's local music library, the VA could execute a search query against an online media catalog with the end goal of streaming the song to the user's device.
Hence, integration of domain knowledge becomes crucial to improve recognition accuracy of such spoken queries.
However, with on-device ASR, offline knowledge sources are constrained because disk space and compute resources are limited. Therefore, integration of knowledge sources, which are often very large and dynamic in nature \cite{VanGysel2020entity}, can be cumbersome.

An empirical analysis of entity-centric information domain queries from a representative sample of anonymized usage logs of a popular VA shows that 36\% of such queries contain a more suitable candidate hypothesis in the \NBestList{} which is not ranked first by the on-device \ac{ASR} system.
One way to overcome the problems associated with recognizing entity-centric queries is to run an on-device domain classifier on top of the on-device \ac{ASR} result, and, for voice commands classified as 
information domain queries, perform server-side \NBest{} rescoring using domain-specific \acp{LM} for knowledge integration.
Since rescoring occurs on server, the domain-specific \acp{LM} are not subject to the same constraints as the on-device models, and therefore can be larger.

Previously, various efforts have been made to improve \ac{ASR} accuracy for entity-rich queries.
\citet{huang2019empirical} conducted an empirical study using Transformer-based \acp{LM} to achieve significant \ac{WER} reductions with second-pass \NBest{} rescoring.
Others \cite{shin2019effective,pelloin2022asr,salazar2020masked,xu2022rescorebert} have shown that using masked \ac{LM} training objectives, like BERT \cite{devlin2019bert}, for \NBest{} rescoring are effective for improving \ac{ASR} accuracy.
\citet{wang2021leveraging} demonstrated the effectiveness of \ac{ASR} \NBest{} information in entity retrieval.
\citet{van2022space} improved \ac{WER} on entity queries by implementing probabilistic grammars as a complement of \NGram{} \ac{LM}s within a \acl{FST} framework.

While the works mentioned above have made significant progress with respect to the use of \acp{LM} for \NBest{} rescoring, they are often limited to considering only a single \ac{LM} architecture at a time and do not consider different subpopulations of the query distribution (that is, head, torso, and tail; see \S\ref{section:experiments:data:stratification}).
Recently, it was shown that different LM architectures lead to better performance on different subsets of the information domain query distribution \cite{van2022space} -- where some architectures work well for head queries, and others perform better on the tail.
In addition, lack of analysis of how an \ac{LM} technique affects different subpopulations of the query distribution may lead to cases where the \ac{ASR} quality degrades on tail queries while the degradation is concealed by overall recognition enhancements, as most of the improvements come from head queries.

In this paper, we investigate strategies for building and combining multiple LMs for \NBest{} rescoring of entity-centric information domain queries. We combine different rescoring LMs and evaluate the recognition quality on different subpopulations (head, torso, and tail \S\ref{section:experiments:data:stratification}) of the information domain query distribution.
Our focus lies on applying \NBest{} rescoring in an application where the on-device ASR system is resource-constrained, while the server-side \NBest{} re-scorers can leverage additional resources and information and hence enhance ASR accuracy.

To the best of our knowledge, our contribution is the first comparison of established techniques for the specific application, extensive empirical evaluation on domain knowledge suitability, using effective model fusion techniques that combine multiple LM architectures with complementary strengths, and testing on different data splits and subpopulations (\S\ref{section:experiments:data:stratification}).

We focus on three \textbf{categories} of LMs and evaluate how signals extracted from each category can contribute to improving \ac{ASR} accuracy. The categories are %
\begin{enumerate*}[label=(\arabic*)]
    \item \NGramCategory[\mathbf]{}: back-off word \NGram{} LMs \cite{Katz1987backoff},
    \item \NNLMCategory[\mathbf]{}: sub-word \acp{NNLM} \cite{Bengio2000nnlm} 
(\S\ref{section:methodology:lms})
, and
    \item \LLMCategory[\mathbf]{}: pretrained \acp{LLM} such as GPT-3 \cite{brown2020language}.
\end{enumerate*}
While \NGramCategory{} and \NNLMCategory{} categories are trained from scratch on domain-specific data, \LLMCategory{} category includes out-of-the-box models accessed as a service. 

With our specific server-side rescoring setting, the inclusion and comparison with the \LLMCategory{} category is meant to serve as an ``out-of-the-box'' baseline that shows the hardness of the problem; i.e., it demonstrates the quality one would be able to achieve by outsourcing the problem to an external \LLMCategory{} service without additional training. Therefore, fine-tuning an \LLMCategory{} is out of the scope of this work. Instead, we focus on whether we are able to construct and assess the ability of various LMs to help improve recognizing domain specific queries, with relatively more controllable modeling scales than \LLMCategory{}s.

We focus on unidirectional LMs since they are more generally applicable (e.g., streaming applications) compared to bidirectional \acp{LM}, and hence, can be repurposed in a variety of settings and applications, as opposed to their bidirectional variants.

We systematically pick representative model architectures within each LM category, and conduct in-depth analysis on an individual category's \NBest{} rescoring results, as well as joint impacts of combining multiple LM categories.
In the end, we compare the single and cross-category \NBest{} rescoring performance for the best rescoring modeling strategy to improve recognition accuracy. 

To this end, our \acp{RQ} are:
\begin{enumerate*}[label=(\textbf{RQ\arabic*})]
	\item Can a single rescoring LM from each LM category reach substantial accuracy improvements on all subpopulations (head, torso and tail) of entity-rich information domain queries?
  \item Is it beneficial to conduct \NBest{} rescoring using domain-specific \acp{LM} trained from scratch, compared with directly using an out-of-the-box external \ac{LLM} service as baseline?
	\item Furthermore, do mixture combinations of multiple \acp{LM} from various categories outperform the single best rescoring \ac{LM}?
\end{enumerate*}

We provide empirical advice on training and selecting \NBest{} rescoring models in a generalizable way.
Our findings are:
\begin{enumerate*}[label=(\arabic*)]
	\item We show that training a single domain-expert \NBest{} rescoring model from \NGramCategory{} category or \NNLMCategory{} category leads to significant \ac{WER} reductions (WERRs) on entity-rich queries across all subpopulations compared to the baseline ASR system.
    \item We find that any of our domain-specific \NGramCategory{} category and \NNLMCategory{} category rescoring model outperforms the out-of-the-box \LLMCategory{} category model baseline significantly, and they are also much smaller in sizes and have fewer numbers of model parameters.
    \item We discover that \NNLMCategory{} category is slightly better than \NGramCategory{} category, but building \NGramCategory{} category models is still beneficial. 
	\item \underline{Most importantly}, we also further discover that effective multi-category rescoring model fusion of \NGramCategory{} and \NNLMCategory{} categories gain complementary advantages over single rescoring models, and results in boosting additional overall accuracy improvements averaging all subpopulations.
\end{enumerate*}

\section{Methodology}
\label{section:methodology}

ASR systems generate ranked \NBest{} lists that consist of multiple candidate hypotheses, by exploring a subset of the entire search space, sorted by decoding signals.

\subsection{N-best rescoring with ASR and server-side LMs}
\label{section:rescoring}

\begin{table}[t]
  \caption{Example of a problematic N-best ranking for an utterance with reference text \emph{``play Dickie Jones movies''} with associated on-device signals (lower is predicted as better). Note that although the best prediction by the ASR system is incorrect, the correct prediction is given at rank 3.\label{tab:nbesterr}}
  \centering\small%
  \renewcommand{\tabcolsep}{5pt}%
  \renewcommand{\arraystretch}{0.75}%
  \begin{tabular}{cccl}
  \toprule
  \textbf{Rank}& \makecell{\textbf{Acoustic}\\\textbf{Signal}}&\makecell{\textbf{LM}\\\textbf{Signal}}&\textbf{Hypothesis}\\
  \midrule
  1 & 208 & 50 & \emph{Play the Key Jones movies} \\
  2 & 286 & 48 & \emph{Play Ricky Jones movies} \\
  3 & 638 & 20 & \emph{Play Dickie Jones movies} \\
  \bottomrule
  \end{tabular}
\end{table}

Table~\ref{tab:nbesterr} shows an example of a problematic \NBest{} list generated by an on-device ASR system, and it suggests room for improvement where the correct hypothesis is included in the top-ranked hypotheses but not ranked at the top.
Rather than generating new hypotheses, we focus on seeking opportunities to optimize the rankings of the hypotheses, also known as \NBest{} reranking.
Comparing to other techniques, it allows us to integrate large, server-side LMs from various resources such as the Davinci model API as described in \S\ref{section:experiments:serverLM}.

In this paper, we obtain multiple signals from on-device ASR and server-side LMs and use linear interpolation to combine the features for integrating domain knowledge to \NBest{} ranking criteria. 
The on-device ASR decoding signals consist of acoustic and LM scores. In the case of a traditional hybrid ASR system \citep[p.~289]{Jurafsky2008slp}, the acoustic and LM signals are provided by the acoustic and language models, respectively. For newer end-to-end systems \citep[\S16.3]{Jurafsky2023slp3draft}, the acoustic signal is provided by the end-to-end model and the LM signal by the external LM of the on-device system.
They enable us to evaluate the costs of the decoded sequences stemmed from acoustic and language models of an \ac{ASR} system. Lower scores indicate a better system.

The additional features, that contribute to the domain knowledge integration ability of the new \NBest{} ranking criteria, are calculated from the server-side LMs.
The features stem from the negative log-likelihood assigned by the LM for each token in a sequence.
To effectively combine the on-device ASR decoding signals (e.g., the acoustic score assigned by the on-device acoustic model) and the server-side LM features, we use a linear model and find the best set of weights by minimizing the \ac{WER} on a validation set using Powell's method \cite{powell1964efficient}, which is effective for non-differentiable cost functions.

\subsection{LM categories under consideration}
\label{section:methodology:lms}

As mentioned in \S\ref{section:introduction}, we consider multiple LM categories.
For the first category (\NGramCategory{}), we consider \NGram{} LM, which is a Markov model where the current token prediction is dependent on a window of history tokens. The conditional probabilities for the tokens stem from the counting in a training corpus. With the development of various smoothing techniques such as Witten-Bell discounting \cite{witten1991zero}, \NGram{} LMs have become one of the classical LMs for speech recognition tasks.

For the second category (\NNLMCategory{}), we consider \ac{FOFE} \cite{zhang2015fixed}, \ac{LSTM} \cite{Hochreiter1997LongSM} and
Transformer~\cite{vaswani2017attention}.
We use sub-word level \acp{NNLM} of varying sizes shown in Table~\ref{tab:nnlm}.

The \ac{FOFE} NNLM is a feed-forward model in which variable-length input sequences are encoded by
fixed-size vectors, with minimal information loss.
Such an architecture has been reported as an accuracy-competitive and performance-efficient language modeling approach~\cite{watcharawittayakul-etal-2018-dual}.

\Ac{LSTM}  is a type of recurrent neural network designed to have better gradient flows. These models have previously been competitive with \NGram{} and Transformer based models when used as a reranking model for ASR systems \cite{Irie2019LanguageMW}. 

Transformer-based architectures achieved state-of-the-art 
results in many language modeling tasks~\cite{devlin2019bert,Irie2019LanguageMW,brown2020language,raffel2020t5}.
The Transformer-based \acp{NNLM} used in this work are built as follows.
Relative positional encoding \cite{shaw2018self} is added to the input embedding vector.
Then, several self-attention encoding blocks are stacked.
We use layer normalization, followed by multi-head attention and residual connections.
A final linear projection with softmax activation is used to determine the subword unit scores.

For the third category (\LLMCategory{}), we consider large language models, which are Transformer decoder-only autoregressive LMs, typically with large number of parameters. Recent developments of the GPT-3.5 series \cite{OpenAI2023GPT35} have shown that they provide high-quality feedback for multiple tasks \cite{brown2020language}, thanks to the parameter sizes and vast amount of training data from a large variety of sources. We are interested to use the GPT-3.5 series for \NBest{} rescoring knowledge integration and expect that the domain knowledge is intrinsic in the GPT-3.5 series.

\section{Experiments}
\label{section:experiments}

\newcommand{\TemplatePrior}{P\left(\text{template}\right)}
\newcommand{\EntityPrior}{P\left(\text{entity}\right)}

\subsection{Entity-heavy query data}
\label{section:experiments:data}

In this paper, we focus on the recognition of entity-centric media player queries. %
As mentioned in \S\ref{section:introduction}, we operate under the setting where ASR runs on-device using a resource-constrained model, and the obtained transcription is then classified to belong to the information domain, thus, requiring access to a knowledge base.
We use the context-free grammar of media player queries published by \citet{van2022space}, and use it to generate media player queries. 
In addition, we generate speech with a 
Neural Text-To-Speech (TTS) system \cite{achanta2021device}
on validation and test splits (\S\ref{section:experiments:data:stratification}) of the generated queries to measure the quality of our \NBest{} rescoring, which
is common practice in many past researches \cite{peyser20_interspeech,huang22j_interspeech,weiran22_interspeech}.
We generate the synthetic validation and test sets mainly because we are constrained by the fact that most of the existing speech recognition test sets do not have good entity coverage. Some directly accessible usage based VA test sets only show 0.83\% entity coverage, which is insufficient. 
Specifically, for the scope of this research, representative entity-rich VA utterances are necessary for investigating the effectiveness of our approach, while most of the available ASR test sets mainly consist of entity-unrelated or general-purpose samples that make it ambiguous whether our approach can demonstrate sufficient evidence for the research questions. Consequently, the synthesis process becomes necessary for effective evaluation in this research.

The query grammar (see \cite{van2022space} for more information) consist of two components:
\begin{enumerate*}[label=(\arabic*)]
\item query templates that contain entity slots and are representative of the VA media player query distribution, each associated with a prior probability, $\TemplatePrior{}$, and
\item a weighted list of entities that can be inserted into the template, with each entity associated with a prior, $\EntityPrior{}$, that correlates with entity popularity.
\end{enumerate*}

To control the complexity of the experiments, we apply cutoffs on the ranked template and entity lists. We keep the top-100 templates according to their prior, and we limit the number of entities to the top-200k. We subsequently sample queries from the joint template/entity probability, $\TemplatePrior{} \cdot \EntityPrior{}$, until each unique query is sampled at least once.

\subsection{Entity query splits and subpopulations}
\label{section:experiments:data:stratification}

\begin{table}
\caption{Statistics of our validation/test set utterances (\S\ref{section:experiments:data:stratification}) sliced by their subpopulations (head/torso/tail). We report the mean ($\mu$) and std. dev. ($\sigma$) of the \NBestList{} lengths. In addition, we also report the best and worst possible WERs by selecting the best and worst hypothesis for each utterance, resp.\label{tab:sets}}
\centering%
\begin{subtable}[t]{0.475\textwidth}
\centering\footnotesize%
\renewcommand{\tabcolsep}{5pt}%
\renewcommand{\arraystretch}{0.75}%
\begin{tabular}{@{}lrrr@{}}%
\toprule%
&\textbf{Head}&\textbf{Torso}&\textbf{Tail}\\%
\midrule%
\# utterances&\numprint{1000}&\numprint{1000}&\numprint{1000}\\%
$\;\;\;\;$ (with N $\ge$ 1)&\numprint{559}&\numprint{596}&\numprint{638}\\%
\NBest{} length ($\mu \pm \sigma$)&2.77 $\pm$ 4.06&3.56 $\pm$ 7.15&3.46 $\pm$ 5.53\\%
Best possible WER&2.05&3.36&3.29\\%
Worst possible WER&10.83&13.33&14.20\\\bottomrule%
\end{tabular}%
\caption{Validation set\label{tab:sets:validate}}%
\end{subtable}
\\
\begin{subtable}[t]{0.475\textwidth}
\centering\footnotesize%
\renewcommand{\tabcolsep}{5pt}%
\renewcommand{\arraystretch}{0.75}%
\begin{tabular}{@{}lrrr@{}}%
\toprule%
&\textbf{Head}&\textbf{Torso}&\textbf{Tail}\\%
\midrule%
\# utterances&\numprint{1000}&\numprint{1000}&\numprint{1000}\\%
$\;\;\;\;$ (with N $\ge$ 1)&\numprint{576}&\numprint{611}&\numprint{622}\\%
\NBest{} length ($\mu \pm \sigma$)&2.88 $\pm$ 4.19&3.20 $\pm$ 4.08&3.70 $\pm$ 10.53\\%
Best possible WER&2.22&2.46&2.90\\%
Worst possible WER & 11.43 & 13.27 & 14.32\\\bottomrule%
\end{tabular}%
\caption{Test set\label{tab:sets:test}}%
\end{subtable}
\end{table}

We randomly split the generated queries into training, validation and test sets, with sampling ratios of 90\%, 5\% and 5\% respectively.
The three sets are disjoint, even though they are sampled from the same carrier phrases and entities distributions, and such sampling from the same distribution technique is an approach widely used in the community including by open evaluations held by NIST. 
For the scope of this work, we do not focus on conducting zero-shot pre-trained LLMs evaluations. Instead, our settings aim to improve accuracy on specific domains that have curated entities from knowledge bases, trending topics, etc and aim for an extensive coverage of them. This is reflected in the fact that we use 200k entities.

The training set is used to train the LM architectures of \NGramCategory{} and \NNLMCategory{} categories (\S\ref{section:methodology:lms}) as follows. %
For the \NGram{} models, the entire training set is used during estimation.
Meanwhile, since the \acp{NNLM} can be expensive to estimate, we take a 500M sample of queries (denoted $\mathbf{T}$) as its training data.

The validation and test splits are partitioned based on the frequency a query occurs in the respective split, with the top 10\% being ``head", 10\% to 50\% being ``torso", and bottom 50\% being ``tail". We subsequently sample 1k queries from each partition, generate audios using TTS (\S\ref{section:experiments:data}), and use the generated audios to obtain \NBest{} lists using our on-device ASR system (\S\ref{section:decoderescore}).
Table~\ref{tab:sets} presents detailed statistics of all subpopulations in the validation and test sets after ASR decoding.

The validation set serves two purposes: 
\begin{enumerate*}[label=(\arabic*)]
 \item to find the optimal linear interpolation weights of the features (\S\ref{section:rescoring}), and
 \item to select sub-word tokenizer and LM hyper-parameters (\S\ref{section:experiments:serverLM}).
\end{enumerate*}

The test sets are only used to report \ac{WER}s obtained by server-side rescoring.

\subsection{On-device ASR system}
\label{section:decoderescore}

Our on-device ASR system uses a \acl{CNN} \acl{AM} similar to \cite{huang2020sndcnn}, a 4-gram word LM in the first  pass and a FOFE word \ac{NNLM} in the second pass.
We generate the \NBest{} lists and decoding signals with the ASR decoder for the validation and test sets. 

\subsection{Server-side LMs for rescoring features}
\label{section:experiments:serverLM}

\paragraph*{Word \NGram{} \acp{LM} (\NGramCategory{}).}
We train back-off \NGram{} models with Witten-Bell smoothing on the training set using SRILM \cite{stolcke2002srilm} for the \NGramCategory{} feature. We sweep the max \NGram{} order over $\{\text{2}, \text{3}, \text{4}\}$ and the pruning threshold over $\{4^{-4}, 4^{-5}, ..., 4^{-19}\} \cup \{\text{0}\}$.

\paragraph*{Sub-word \acp{NNLM} (\NNLMCategory{}).}
Before training NNLMs, we first conduct \ac{SP} modeling \cite{kudo2018sentencepiece} and encode the query texts in $\mathbf{T}$ with the \ac{SP} model,
which is expected to be helpful in handling rare words \cite{huang2019empirical} for entity recognition tasks.
We determine the \ac{SP} vocabulary size through a pilot study where we select the optimal size on the validation set (\S\ref{tab:sets:validate}) by sweeping the vocabulary size over $\{\text{15k}, \text{36k}, \text{48k}\}$ and we select an optimal size of 15k. 
Various \ac{NNLM} architectures (\S\ref{section:methodology:lms}) are trained on \ac{SP}-encoded $\mathbf{T}$ as per the hyper-parameter configurations outlined in Table~\ref{tab:nnlm}.
We select these configurations because we find that bigger NNLMs trained on $\mathbf{T}$ with more parameters than the ones shown in Table~\ref{tab:nnlm} present undesirable performance on validation sets because of overfitting.

\begin{table}[t]
  \caption{Hyper-parameters of server-side \acp{NNLM}.\label{tab:nnlm}}
  \centering\footnotesize%
  \renewcommand{\tabcolsep}{5pt}%
  \renewcommand{\arraystretch}{0.75}%
  \begin{tabular}{
   l c c c c c
   *{6}{S[table-format=3.0]}
  }
  \toprule
 \textbf{Model} & \textbf{Param.} & \textbf{Layers} & \thead{\textbf{Embed.} \\ \textbf{dim}} & \thead{\textbf{Layers} \\ \textbf{dim}} & \thead{\textbf{Att.} \\ \textbf{heads}}\\
  \midrule
  LSTM
  &  28MM & 2 &  512 &  768 & - \\
  & 108MM & 4 &  896 & 1536 & - \\
  & 217MM & 5 & 1536 & 2048 & - \\
  \midrule
  FOFE
  &  31MM &  8 &  1024 & 1024 &-\\
  &110MM & 12 & 2048 & 2048 &-\\
  &447MM & 16 & 4096 & 4096 &-\\
  \midrule
  Transformer
  &  30MM &  6 &  512 &  512 & 8 \\
  & 114MM & 12 &  896 &  896 & 14 \\
  & 450MM & 19 & 1536 & 1536 & 24 \\
  \bottomrule
  \end{tabular}
\end{table}

The models of \NNLMCategory{} category are trained with the Adam optimizer~\cite{Kingma2014Adam} on 16 GPUs for 80 epochs, each with 28k training steps and 16 sequences per minibatch.
A warmup stage runs to linearly increase the learning rate between $10^{-6}$ and $10^{-3}$ for 1.2k steps, and subsequently the learning rate decreases exponentially with a factor $0.94$.
Dropout with a fixed rate 0.1 is applied.
Internal layers use RELU activations.
The \ac{FOFE} models have a \ac{FOFE} factor of 0.85 and a \ac{FOFE} order of 8.

At inference time, we tokenize an \NBest{} candidate using the \ac{SP} model and subsequently compute the joint log-likelihood using the trained NNLMs for the \NNLMCategory{} feature.

\paragraph*{Davinci (\LLMCategory{}).}
The latest Davinci \cite{OpenAI2023Davinci} models build upon InstructGPT \cite{ouyang2022training} and \emph{text-davinci-003} is a reinforcement learning with human feedback (RLHF) \cite{christiano2017deep} model of about 175 billion parameters that improves the previous model series. We directly use OpenAI API with Davinci models for the \LLMCategory{} feature by including token log-probabilities in the API return and then calculate the joint log-likelihood of the corresponding \NBest{} candidate.  

\paragraph*{Model combination.}
We combine the server-side features from the trained \NGramCategory{}, \NNLMCategory{} and \LLMCategory{} category LMs with the on-device ASR signals by linear interpolation with coefficients estimated by Powell's Method as described in (\S\ref{section:rescoring}) for \NBest{} rescoring and domain knowledge integration, and then evaluate the validation and test rescoring WERs on different subpopulations (head, torso and tail).

\section{Results and Discussions}
\label{section:results}

\begin{table}[t]
\caption{Test set WERs for the best single and multiple model combinations are presented, with corresponding relative improvements in the parentheses. Overall best test sets WERs are shown \textbf{bold} and the in-group best WERs are \underline{underlined}. The best model architectures and combinations are selected based on the best validation sets accuracy. We only specify the best model architecture once per model category (R3--R6), but use the same configuration consistently through our experiments.\label{table:wer}}
\centering\footnotesize%
\renewcommand{\tabcolsep}{5pt}%
\renewcommand{\arraystretch}{0.75}%
\begin{tabular}{@{}lrrrr@{}}%
\toprule%
&\textbf{Head}&\textbf{Torso}&\textbf{Tail}&\textbf{Avg.}\\%
\midrule%
R1: On-device signals only&\makecell{4.26\\\phantom{{\scriptsize(0.00\%)}}}&\makecell{5.31\\\phantom{{\scriptsize(0.00\%)}}}&\makecell{5.70\\\phantom{{\scriptsize(0.00\%)}}}&\makecell{5.09\\\phantom{{\scriptsize(0.00\%)}}}\\%
\midrule%
\multicolumn{5}{@{}l}{\textbf{On{-}device signals + (single server-side LM)}}\\%
\midrule%
R2: Davinci ($\sim$175B)&\makecell{4.00\\{\scriptsize(6.10\%)}}&\makecell{5.12\\{\scriptsize(3.58\%)}}&\makecell{5.28\\{\scriptsize(7.37\%)}}&\makecell{4.80\\{\scriptsize(5.70\%)}}\\%
R3: \NGram{} (32.9M)&\makecell{3.40\\{\scriptsize(20.19\%)}}&\makecell{3.73\\{\scriptsize(29.76\%)}}&\makecell{3.92\\{\scriptsize(31.23\%)}}&\makecell{3.68\\{\scriptsize(27.64\%)}}\\%
R4: LSTM (108M)&\makecell{3.38\\{\scriptsize(20.66\%)}}&\makecell{3.65\\{\scriptsize(31.26\%)}}&\makecell{\textbf{\underline{3.89}}\\{\scriptsize(31.75\%)}}&\makecell{3.64\\{\scriptsize(28.49\%)}}\\%
R5: FOFE (110M)&\makecell{\textbf{\underline{3.19}}\\{\scriptsize(25.12\%)}}&\makecell{3.67\\{\scriptsize(30.89\%)}}&\makecell{4.22\\{\scriptsize(25.96\%)}}&\makecell{3.69\\{\scriptsize(27.44\%)}}\\%
R6: Transformer (114M)&\makecell{3.28\\{\scriptsize(23.00\%)}}&\makecell{\underline{3.48}\\{\scriptsize(34.46\%)}}&\makecell{4.00\\{\scriptsize(29.82\%)}}&\makecell{\underline{3.59}\\{\scriptsize(29.54\%)}}\\%
\midrule%
\multicolumn{5}{@{}l}{\textbf{On{-}device signals + LSTM LM +}}\\%
\midrule%
R7: \NGram{}&\makecell{\underline{3.28}\\{\scriptsize(23.00\%)}}&\makecell{\underline{3.83}\\{\scriptsize(27.87\%)}}&\makecell{4.17\\{\scriptsize(26.84\%)}}&\makecell{\underline{3.76}\\{\scriptsize(26.13\%)}}\\%
R8: \NGram{} + Davinci&\makecell{3.28\\{\scriptsize(23.00\%)}}&\makecell{3.84\\{\scriptsize(27.68\%)}}&\makecell{\underline{4.16}\\{\scriptsize(27.02\%)}}&\makecell{3.76\\{\scriptsize(26.13\%)}}\\%
\midrule%
\multicolumn{5}{@{}l}{\textbf{On{-}device signals + FOFE LM +}}\\%
\midrule%
R9: \NGram{}&\makecell{\underline{3.24}\\{\scriptsize(23.94\%)}}&\makecell{\underline{3.70}\\{\scriptsize(30.32\%)}}&\makecell{\underline{4.36}\\{\scriptsize(23.51\%)}}&\makecell{\underline{3.77}\\{\scriptsize(26.00\%)}}\\%
R10: \NGram{} + Davinci&\makecell{3.24\\{\scriptsize(23.94\%)}}&\makecell{3.70\\{\scriptsize(30.32\%)}}&\makecell{4.36\\{\scriptsize(23.51\%)}}&\makecell{3.77\\{\scriptsize(26.00\%)}}\\%
\midrule%
\multicolumn{5}{@{}l}{\textbf{On{-}device signals + Transformer LM +}}\\%
\midrule%
R11: \NGram{}&\makecell{\underline{3.29}\\{\scriptsize(22.77\%)}}&\makecell{\textbf{\underline{3.46}}\\{\scriptsize(34.84\%)}}&\makecell{\underline{3.92}\\{\scriptsize(31.23\%)}}&\makecell{\textbf{\underline{3.56}}\\{\scriptsize(30.12\%)}}\\%
R12: \NGram{} + Davinci&\makecell{3.31\\{\scriptsize(22.30\%)}}&\makecell{3.46\\{\scriptsize(34.84\%)}}&\makecell{3.94\\{\scriptsize(30.88\%)}}&\makecell{3.57\\{\scriptsize(29.86\%)}}\\\bottomrule%
\end{tabular}%
\end{table}

Table~\ref{table:wer} shows rescoring \acp{WER} for the various LM categories where $R1$ corresponds to the on-device ASR system only, $R2$ to the addition of the out-of-the-box ``Davinci" model (\LLMCategory{} category), and $R3$ to the usage of \NGram{} model (\NGramCategory{} category) for rescoring. $R4$, $R5$, $R6$ correspond to the inclusion of LSTM, FOFE and Transformer server-side LMs (\NNLMCategory{} category) respectively. For each LM in $R2$-$R6$, we report the number of parameters of the corresponding model after picking the best model hyper-parameters on the validation set (\S\ref{section:experiments:serverLM}).

For \textbf{(RQ1)}, we use on-device ASR ($R1$) as the baseline. 
As shown in $R3$ to $R6$, comparing to baseline $R1$, all average WERs (column $\textbf{Avg.}$) are significantly better, between 27\%-30\% relative.
\NBest{} rescoring with an \NGramCategory{} category model ($R3$) demonstrates substantial 28\% average WER reduction, 
but we can achieve more significant improvement when rescoring with \NNLMCategory{} category models, such as Transformer ($R6$), which leads to accuracy improvement of over 23\% on head, 34\% on torso, 30\% on tail, and 30\% on average. 
Therefore, our answer to \textbf{(RQ1)} is in the affirmative: integrating a single server-side domain-expert LM is indeed effective for optimizing \NBest{} rescoring and improving entity recognition accuracy compared to on-device ASR only.

Moreover, for \textbf{(RQ2)}, we compare with the out-of-the-box baseline of GPT-3.5 ``Davinci" ($R2$) rescoring results.
By comparing $R3$-$R6$ to $R2$, we also observe that the \NGram{} model
(\NGramCategory{} category) and the sub-word server-side LMs (\NNLMCategory{} category) trained from domain-specific data from scratch outperform the out-of-the-box baseline of GPT-3.5 ``Davinci"
model (\LLMCategory{} category).
For example, by comparing $R6$ with $R2$, rescoring with Transformer achieves substantial relative accuracy enhancements of 18\% on head, 32\% on torso, and 24\% on tail over the GPT-3.5 ``Davinci" rescoring WERs. 
Therefore, our answer to \textbf{(RQ2)} is also positive. It is beneficial to take advantage of training models from \NGramCategory{} category and \NNLMCategory{} category on domain-specific data with decent numbers of model parameters, rather than outsourcing directly to an out-of-the-box \LLMCategory{} category model for effective \NBest{} rescoring.

In terms of \textbf{(RQ3)}, we compare the WERs after rescoring \NBest{} with the interpolation results (\S\ref{section:rescoring}) of multiple server-side LMs from different categories in Table~\ref{table:wer} ($R7$-$R12$).
Inspired by the results obtained as part of answering \textbf{(RQ1)}, we are motivated to investigate whether the combination of multiple LM categories can further improve recognition quality.
Furthermore, because \NGramCategory{} category and \NNLMCategory{} category outperform \LLMCategory{} category ($R3$-$R6$ over $R2$) as shown in the part of answering \textbf{(RQ2)}, we assess multiple \NGramCategory{} category and \NNLMCategory{} category combinations as shown in (\{$R7$, $R9$, $R11$\}). For completeness, we also conduct all \{\NGramCategory{}, \NNLMCategory{}, \LLMCategory{}\} category combinations with results shown in (\{$R8$, $R10$, $R12$\}). 
In general, by observing $R7$ to $R12$ in Table~\ref{table:wer}, we draw the following conclusions:
\begin{enumerate}[label=(\textbf{\arabic*})]
    \item Integrating an \NGramCategory{} category model with a sub-word \NNLMCategory{} category model (Transformer) leads to the best average WER improvement ($R11$) among all of our experiments.
    \item  More specifically, combination of \NGramCategory{} and \NNLMCategory{} category rescoring LMs ($R11$) shows statistically significant (Student's t-test $p$-value $< 0.03$) tail-entity WER improvement over a single Transformer rescoring model ($R6$), best-of-all on torso-entity accuracy among  all of our experiments, and head-entity WER improvement over a single \NGram{} rescoring model ($R3$).
    \item Introducing additional \LLMCategory{} category models does not bring extra benefit for accuracy enhancement. Therefore, the dominating factors for the significant \NBest{} rescoring improvements are the \NGramCategory{} and \NNLMCategory{} category rescoring LMs. 
    \item One trade-off is that the best average WER from $R11$ does not always guarantee the best WERs on each subpopulation individually among our experiments. However, we consider the trade-off acceptable.
\end{enumerate}
Therefore, multi-model combinations from different model categories for \NBest{} rescoring gain complementary advantages of improving head, torso and tail subpopulations respectively over single rescoring models. By joining the multiple strengths on each head/torso/tail subpopulations from each single rescoring LMs in an effective way by model fusion described in \S\ref{section:rescoring}, we are able to collectively reach optimal WERs for each subpopulation and consequently the best average WERR over all head/torso/tail sets among our experiments.
In conclusion, our answer to \textbf{(RQ3)} is also positive: we suggest to combine multiple rescoring LMs from different categories to further improve the \NBest{} rescoring accuracy, since combinations of various LM categories outperform use of individual models only.

\section{Conclusions}
\label{section:conclusions}

We showed that training domain-specific server-side LMs for \NBest{} rescoring led to significant accuracy improvements on all head, torso and tail subpopulations. 
We focused on three LM categories and investigated modeling strategies for \NBest{} rescoring.
Training sub-word \ac{NNLM}s (\NNLMCategory{} category) on domain-centric data with a \acl{SP} tokenizer was the most effective single rescoring modeling choice.
Using a single sub-word NNLM, we improved accuracy by 30\% and 34\% for difficult tail and torso entities and 23\% for head entities.
The best single sub-word NNLM trained from scratch also outperformed the out-of-the-box pretrained large LM baseline (\LLMCategory{} category) significantly by 25\% averaging all subpopulations. 

Furthermore, integrating multiple server-side LMs of different categories for rescoring led to additional accuracy improvements over single rescoring LMs and consequently the best WERRs among all of our experiments, thanks to the effective 
model fusion with interpolation coefficients estimated by Powell's method on the validation sets, which combined the complementary advantages and strengths of multiple single rescoring LMs together for \NBest{} rescoring (\S\ref{section:rescoring}), and consequently improved the ASR accuracy to a great extent on all entity-heavy head/torso/tail subpopulations in our experiments.
By combining sub-word \acp{NNLM} (\NNLMCategory{} category) and \NGram{} model (\NGramCategory{} category), we further reached substantial average \ac{WER} reduction of 30\% relative across all subpopulations.

In conclusion, training multiple \NBest{} rescoring models of various categories on domain-specific data, integrating information domain knowledge based on server-side rescoring LMs with individual strengths, and boosting \NBest{} accuracy with complementary signals by model fusion, led to the effectiveness in enhancing entity speech recognition accuracy and improvements on on-device VA systems.

As for future work, we plan to include more LM categories such as autoencoding LMs, conduct fine-tuning on pretrained models, as well as knowledge distillation with teacher-student learning. We also plan to expand varieties of domain data.

\bibliographystyle{IEEEtranN}
\bibliography{arxiv2023-entity_models}

\end{document}